\documentclass[conference]{IEEEtran}
\IEEEoverridecommandlockouts
\usepackage{cite}
\usepackage{amsmath,amssymb,amsfonts}
\usepackage{algorithmic}
\usepackage{graphicx}
\usepackage{multirow}
\usepackage{comment}
\usepackage{textcomp}
\usepackage{xcolor}
\usepackage{amssymb}
\usepackage{pifont}
\def\BibTeX{{\rm B\kern-.05em{\sc i\kern-.025em b}\kern-.08em
    T\kern-.1667em\lower.7ex\hbox{E}\kern-.125emX}}
\begin{document}

\title{A Scalable  Multilabel Classification to Deploy Deep Learning Architectures For Edge Devices 
}

\author{\IEEEauthorblockN{1\textsuperscript{st} Tolulope A. Odetola}
\IEEEauthorblockA{\textit{Electrical and Computer Engineering} \\
\textit{Tennessee Technological University}\\
Cookeville TN 38501, USA \\
taodetola42@students.tntech.edu}
\and
\IEEEauthorblockN{2\textsuperscript{nd} Ogheneuriri Oderhohwo}
\IEEEauthorblockA{\textit{Electrical and Computer Engineering} \\
\textit{Tennessee Technological University}\\
Cookeville TN 38501, USA \\
ododerhohw42@students.tntech.edu}
\and
\IEEEauthorblockN{ Syed Rafay Hasan}
\IEEEauthorblockA{\textit{Electrical and Computer Engineering} \\
\textit{Tennessee Technological University}\\
Cookeville TN 38501, USA\\
shasan@tntech.edu}
}

\maketitle

\begin{abstract}
Convolution Neural Networks (CNN) have performed well in many applications such as object detection, pattern recognition, video surveillance and so on.  CNN carryout feature extraction on labelled data to perform classification. Multi-label classification assigns more than one label to a particular data sample in a data set. In multi-label classification, properties of a data point that are considered to be mutually exclusive are classified. However, existing multi-label classification requires some form of data pre-processing that involves image training data cropping or image tiling.  The computation and memory requirement of these multi-label CNN models makes their deployment on edge devices (IoT devices) challenging. In this paper, we propose a methodology that solves this problem by extending the capability of existing multi-label classification and provide models with lower latency that requires smaller memory size when deployed on edge devices. We make use of a single CNN model designed with multiple loss layers and multiple accuracy layers. The number of accuracy and loss layers are based on number of mutually exclusive attributes to be classified by the multi-label classification. This methodology is tested on state-of-the-art deep learning algorithms such as AlexNet, GoogleNet and SqueezeNet using the Stanford Cars Dataset and deployed on Raspberry Pi3. From the results the proposed methodology achieves comparable accuracy with 1.8x less MACC operation, 0.97x reduction in latency and 0.5x, 0.84x and 0.97x reduction in size for the generated AlexNet, GoogleNet and SqueezeNet CNN models respectively when compared to conventional ways of achieving multi-label classification like hard-coding multi-label instances into single labels. The methodology also yields CNN models that achieve 50\% less MACC operations, 50\% reduction in latency and size of generated versions of AlexNet, GoogleNet and SqueezeNet respectively when compared to conventional ways using 2 different single-labelled models to achieve multi-label classification.
\end{abstract}

\begin{IEEEkeywords}
Multilabel classification, deep learning, transfer learning, Convolution Neural Networks (CNN)
\end{IEEEkeywords}

\section{Introduction}
Convolution Neural Networks (CNN) models have achieved significant success in machine learning and have found adoption in many fields \cite{tolu1}. CNNs are specially successful in computer vision \cite{hailesellasie2019mulnet}. In many computer vision related applications, a single image may be associated with two or more mutually exclusive properties that can be represented with different labels. The representation of a data point with multiple mutually exclusive labels is called Multi-label Classification \cite{scikit}. One application of multi-label classification can be illustrated when considering security, privacy~\cite{baza2019b,park1,baza2018blockchain,baza2019blockchain,baza2019detecting,Park2019,pazos2019privacy,baza2015efficient} in different applications~\cite{baza10,baza2,baza3,baza8,baza4,baza9} or safety critical establishments where facial recognition applications are employed to detect and classify faces to determine access of an individual to secure zones. Multi-label classifier offers the advantage of designing and utilizing one model to perform multiple classifications on single data instance of such data set of images. Multi-label classification has also been demonstrated in text categorization, urban planning and bio-informatics \cite{young2018recent}. 

Several approaches to achieve effective and efficient multilabel classification have been adopted over the years. Zeggada et. al in\cite{ zeggada2016multilabel}  proposes an approach to investigate the behavior of a CNN as a multilabel classifier by utilizing  Radial Basis Function Neural Networks (RBFNN). In their approach, the input images from Unmanned Aerial Vehicles (UAV) are subdivided into equal tiles. Each tile represents each label. Feature extraction is performed on each tile and is used to train a deep learning model. Zhang et. al \cite{ zhang2018multilabel} proposes a Regional Latent Semantic Dependencies (RLSD) model to achieve multilabel classification. This approach leverages on region-based feature extraction and Recurrent Neural Network (RNN) to capture the latent semantic dependencies at the regional level of an input image and classify them on different labels to perform multilabel classification. In reference \cite{ zhang2018convolutional}, Zhang et. al proposes a multilabel classification approach to detect printed circuit boards (PCB) defects. In their approach, PCB images are cropped. These cropped images represent different labels. Feature extraction is carried out on these cropped images and are used for classification.

One shortcoming in these approaches is the problem of scalability attached to the cost of performing multi-labelling of the large data samples. Moreover, these approaches require further image processing such as image tiling and cropping where each tile or cropped image is used to represent different labels of the image.

In this paper, we propose a method of multilabel classification methodology that extends the capability of a model trained for conventional classification for multilabel classification. Our methodology provides a means of achieving multilabel classification with little cost of multi-labelling and achieves high accuracy. The proposed approach also allows the utilization of data fusion techniques to train a model that can classify related and unrelated data samples simultaneously. In this paper, we leverage on transfer learning to modify popular deep learning models.  Our contribution are listed as follows:

\begin{figure*}[t]
\centering
\includegraphics[width=.88\linewidth]{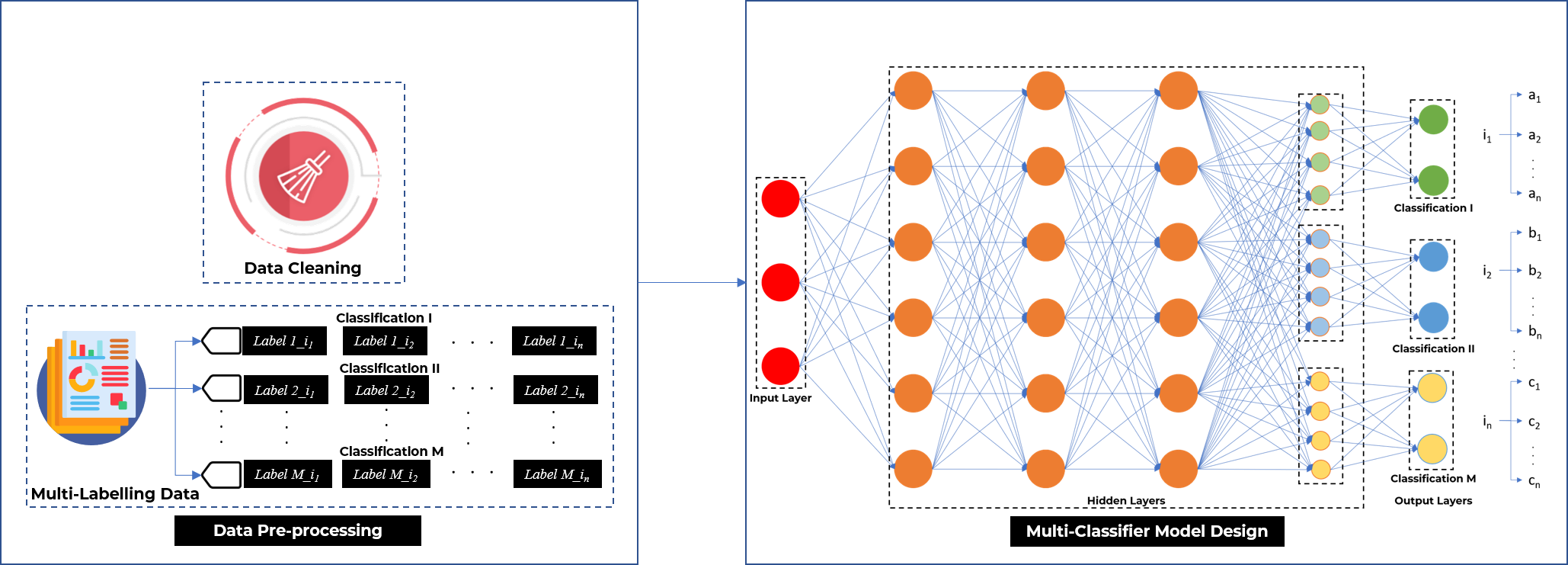}
\caption{Proposed Methodology}
\label{fig_meth}
\end{figure*}

\begin{itemize}
\item A scalable methodology for designing multi-label classification models without the need of image pre-processing like image cropping or tiling
\item An approach to designing multi-label classification models that achieves faster latency, reduction in MACC (multiply and accumulate) operations and reduction in the size when deployed on edge devices when compared with more than one single labelled models or hard-coding multiple label instances into a single label to achieve multi-label classification.
\end{itemize}

The remainder of this paper is organized as follows: Section II discusses the proposed methodology. Section III describes the methodology implementation. Section IV discusses our results. Section V shows the comparison with state-of-the-art multilabel methods. Section VI discusses other related works and Section VII concludes the paper.

\begin{figure*}[t]
\centering
\includegraphics[width=.88\linewidth]{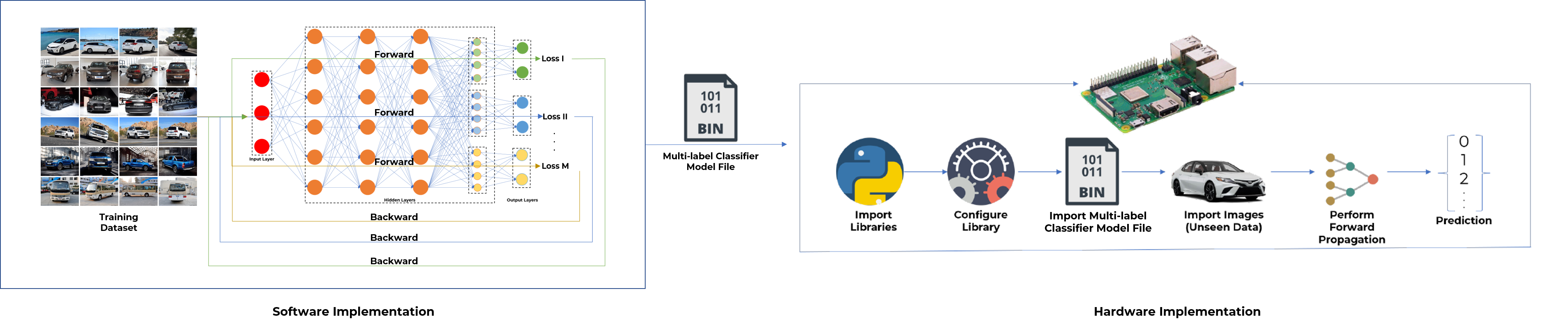}
\caption{Software/Hardware Implementation of Proposed Methodology}
\label{fig_impl}
\end{figure*}

\begin{figure*}[!t]
\centering
\includegraphics[width=.88\linewidth]{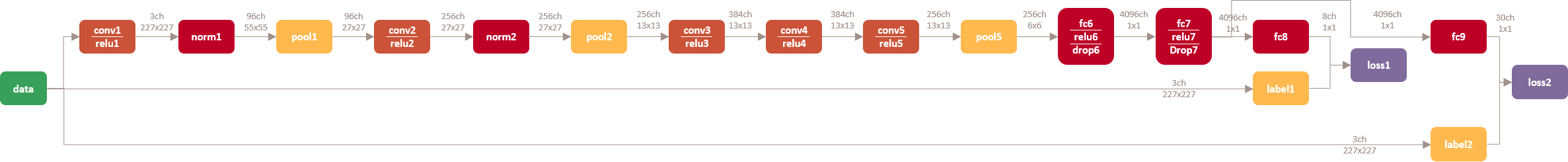}
\caption{Netscope Generated for AlexNet using the Proposed Method}
\label{fig_alex}
\end{figure*}

\begin{figure}[!h]
\centering
\includegraphics[width=.88\linewidth]{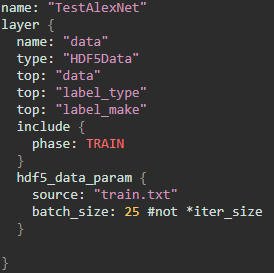}
\caption{Data Layer for Caffe Model Training}
\label{data}
\end{figure}

\section{Proposed Methodology}
The proposed methodology shown in Fig. \ref{fig_meth} is used to perform multi-label classification on the Stanford car datasets. The goal of this methodology is to design a multi-class deep learning model that can perform multiple classification on an input data instance to concurrently predict more than one mutually exclusive properties during forward propagation.  The methodology is elaborated as follows:

\subsection{Data Pre-processing}
\label{AA}
Prior to training, each training data instance is labelled based on the number of categories (number of mutually exclusive attributes to be classified) of classification required by the model as shown in the data pre-processing block in Fig \ref{fig_meth}. In literature, for multi-label classification, to extract mutually exclusive properties image cropping or tiling is employed to single out features of each respective attributed to be classified. RNNs have also been employed to perform multiple feature extraction for each image. These techniques increases data pre-processing overhead cost for huge amount of data.

In our proposed methodology, a training image data instance has different number of labels based on the number of mutually exclusive attributes to be classified as shown in Fig. \ref{fig_meth}. To illustrate, for a image data set $x_{d}$ of size $d$ where each image data instance $x_i$ belongs to $n$ number of label categories where each category represent the number of mutually exclusive properties of the image data instance to be classified. Each training image data instance is assigned $n$ number of labels 
such that: ($x_{i}$, $i_1$, $i_2$, ... $i_n$) 

Hence:
\begin{align}
\{{1_{i1}, 2_{i1}, ..., M_{i1}}\} \in i_1\\
\{{1_{i2}, 2_{i2}, ..., M_{i2}}\} \in i_2\\
\{{1_{in}, 2_{in}, ..., M_{in}}\} \in i_n
\end{align}

 Where:

$x_i$ represents a data instance from the image data set $x_d$

$i_1$, $i_2$, ... $i_n$ represents the label categories for the same data but each label category contains different labels corresponding to the number of mutually exclusive attributes to be classified.

$M_{i1}$, $M_{i2}$, ..., $M_{in}$ represent the number of classes in each respective label category of $i_1$, $i_2$, ... $i_n$

\subsection{Model Design}
The Multi-Classifier Model Design block in Fig. \ref{fig_meth} summarizes our approach of designing a multi-label CNN model. In this methodology, the different labels for each category are referenced. Hence, there are different loss and accuracy layers which matches the number of label categories.


At the output declaration stage, classification layers (fully connected layers) are assigned to the various label categories in order to perform multi-classification as illustrated in the different sets of classification in Fig. \ref{fig_meth}. The respective classification layers for each label category contain neurons corresponding to the number of sub-classes in each respective label category. 

The model performs $n$ number of predictions that tallies with the number of label categories, losses and accuracy as depicted in the software implementation section of Fig. \ref{fig_impl}. For each classification category, an accuracy and a loss value is generated respectively. This allows for multiple backward propagations and weight updates for corresponing to the number of losses and consequently the number of label categories.

\section{Methodology Implementation}
The implementation of the proposed methodology is in 2 folds namely:
\begin{itemize}
    \item Software Implementation
    \item Hardware Implementation
\end{itemize}
\subsection{Software Implementation}
To demonstrate this methodology, Caffe \cite{jia2014caffe} deep learning framework is used as shown in the model training block Fig. \ref{fig_impl}. The methodology was demonstrated on state-of-the-art deep learning models like AlexNet, SqueezeNet and GoogleNet leveraging on transfer learning. These models are used to train Standford car dataset respectively. The Standford car dataset containing 16000 training instance is processed such that each training instance possesses 2 label categories namely: car type and car make. Each training data instance has 2 labels following the data pre-processing methodology as shown in Fig. \ref{data}.

\begin{figure}[!h]
    \centering
    \includegraphics[width=.6\linewidth]{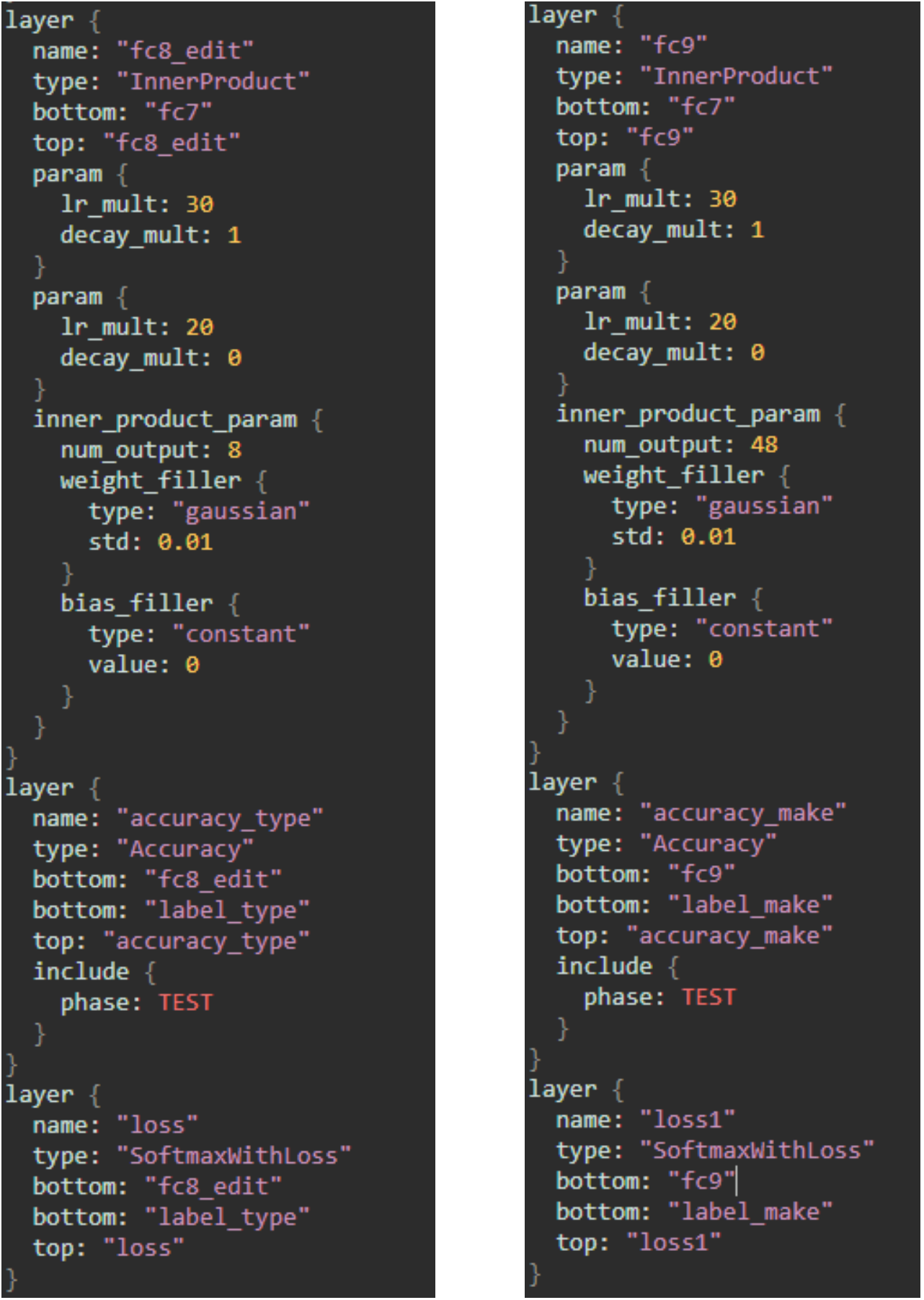}
    \caption{Fully Connected Layer Configuration for Classification Layers for Car Type and Car Make}
    \label{fc}
\end{figure}

The model contains 2 classification layers, 2 loss layers and 2 accuracy layers as shown if Fig. \ref{fc} that tallies with the number of label categories.

Transfer learning is employed for the training models due to the size of the dataset. With transfer learning, we make use of  using large model architecture while leveraging on the pre-trained weights. Fig. \ref{fig_alex} shows the full architecture of AlexNet when the proposed methodology is adopted. During the training, all layers are frozen except the last 2 classification layers used for the multi-label classification. After training is complete, a multi-label classification model file is generated.

\subsection{Hardware Implementation}
The process flow for the deployment of the CNN model on an IoT or edge device is shown in Fig. \ref{fig_impl}. For the multi-label classifier inference, depending on the technology relevant Python libraries are installed and imported on the IoT device. After installation and successful importation of the libraries, the multi-label model is deployed on the IoT device to perform multiple forward propagation simultaneously on input images based on the number of mutually exclusive properties defined in the label categories to the classified. Test image datasets are passed through the deployed model on the IoT device to determine the accuracy, latency and throughput. The memory and computation requirement on the IoT device are also evaluated and compared with existing methodologies of designing CNN multi-label models.

%

\section{Results and Discussion}
The multi-label methodology is carried out on 3 state-of-the-art CNN models namely: AlexNet, SqueezeNet and GoogleNet. These CNN models are trained on Stanford car datasets leveraging on transfer learning. The datasets consist of 16000 training data. 


Each training data instance has 2 label categories consisting of car make ( for instance Toyota, Nissan, etc.) and car type ( for instance SUV, Coupe, etc.). The car make label category consist of 48 classes while the car type label consist of 8 classes. This methodology is compared with conventional deep learning techniques used to achieve multi-label classification. The conventional multi-label technique used to achieve comparison include hard-coding (HC) the multiple labels to form one label such that every data instance have one label category instead of 2 (for instance Toyota-SUV, Nissan-Coupe, etc.). This approach of labelling is seen in ImageNet \cite{russakovsky2015imagenet} training data. Another technique adopted for comparison is the 2 model (2M) approach. Here each data instance are trained with a single label respectively. 2 CNN models are designed trained to perform classification for car make and car type respectively. Training of the CNN models is done using Caffe deep learning framework and they are deployed on Raspberry Pi3. 100 test data samples are used to perform classification in order to obtain their their latency.

Tables \ref{tab_alex}, \ref{tab_goog} and \ref{tab_squeeze} show comparisons between the proposed multi-label approach, the HC method and the 2M method using AlexNet, SqueezeNet and GoogleNet. 

Table \ref{tab_alex} shows the comparison between the 3 techniques when applied on AlexNet. From Table \ref{tab_alex}, it is observed that the proposed approach shows approximately 50\% decrease in terms of the amount of multiply and accumulate (MACC), in terms of the size of the model and also in terms of latency when compared to the 2M approach. The accuracy and loss values of the 2 approaches are similar and fall within the close regions.

The HC approach for the dataset results in 108 classes based on the available data. The HC approach is very restrictive and is void of robustness when compared to the proposed approach. The proposed approach has 2 classification layers that individually caters for the 48 classes for the car make category and the 8 classes for the car type category simultaneously. Hence despite the limited data, the proposed approach can cater for 384 classes if the label categories are merged together compared to the 108 classes of the HC technique.


\begin{table}[!h]
\caption{Comparison of Approaches Applied to AlexNet CNN Model}
\begin{tabular}{|c|c|c|c|c|}
\hline
                             &         & Proposed Approach     & 2M         & HC       \\ \hline
\multirow{2}{4em}{Accuracy}    & Make    & 0.55                  & 0.64                & 0.61 \\ \cline{2-4}
                             & Type    & 0.72                  & 0.43                  \\ \hline

\multirow{2}{4em}{Loss}        & Make    & 1.57                  & 0.98                  & 1.62 \\ \cline{2-4}
                             & Type    & 0.83                  & 1.99                  &                       \\ \hline
\multicolumn{2}{|c|}{Size (MB)}             &223.00                       &445.20                       &445.2                       \\ \hline
\multicolumn{2}{|c|}{\# of Parameters} & 20.36M                & 40.50M                & 20.33M                \\ \hline
\multicolumn{2}{|c|}{\# of Trained Layer MACC}       & 17.00M               & 33.78M               & 17.21M               \\ \hline
\multicolumn{2}{|c|}{Latency (secs.)}          &17.215     &35.498      &17.693  \\ \hline
\end{tabular}
\label{tab_alex}
\end{table}

\begin{table}[!h]
\caption{Comparison of Approaches Applied to GoogleNet CNN Model}
\centering
\begin{tabular}{|c|c|c|c|c|}
\hline
                             &         & Proposed Approach     & 2M         & HC       \\ \hline
\multirow{2}{*}{Accuracy}    & Make    & 0.75                  & 0.84                  & 0.83 \\ \cline{2-4}
                             & Type    & 0.99                  & 0.81                  &                       \\ \hline
\multirow{2}{*}{Loss}        & Make    & 0.86                  & 1.15                  1.29 \\ \cline{2-4}
                             & Type    & 0.51                  & 1.21                  &                       \\ \hline
\multicolumn{2}{|c|}{Size (MB)}             &40.9                       &95.6                       &48.6                       \\ \hline
\multicolumn{2}{|c|}{\# of Parameters} & 8.28M                & 18.36M                & 9.92M                \\ \hline
\multicolumn{2}{|c|}{\# of Trained Layer MACC}       & 57.344K              &57.344K                & 107.520K              \\ \hline
\multicolumn{2}{|c|}{Latency (secs.)}          &3.178  &7.265  &3.277  \\ \hline
\end{tabular}
\label{tab_goog}
\end{table}

\begin{table}[h]
\caption{Comparison of Approaches Applied to SqueezeNet CNN Model}
\begin{tabular}{|c|c|c|c|c|}
\hline
                             &         & Proposed Approach     & 2M         & HC       \\ \hline
\multirow{2}{*}{Accuracy}    & Make    & 0.65                  & 0.75                  & 0.66 \\ \cline{2-4}
                             & Type    & 0.74                  & 0.70                 &                       \\ \hline
\multirow{2}{*}{Loss}        & Make    & 1.47                  & 1.76                  & 2.84 \\ \cline{2-4}
                             & Type    & 0.89                  & 2.40                  &                       \\ \hline
\multicolumn{2}{|c|}{Size (MB)}             &2.9                       &5.7                       &3                       \\ \hline
\multicolumn{2}{|c|}{\# of Parameters} & 749.5k                & 1470.26k               & 745.39k                \\ \hline
\multicolumn{2}{|c|}{\# of Trained Layer MACC}       & 28.728K               & 28.728K               & 53.865K               \\ \hline
\multicolumn{2}{|c|}{Latency (secs.)}          &6.466   &12.848   &6.512   \\ \hline
\end{tabular}
\label{tab_squeeze}
\end{table}

For AlexNet from Table \ref{tab_alex}, it is seen that the accuracy of the proposed approach lies in close region when compared with the accuracy of the HC technique. The proposed method shows a lower loss when compared with the HC method. The proposed method can cater of 384 classes while using 60.87\% less MACC compared to the HC method which caters for just 108 classes. The size of the model produced by the proposed method is are comparable with the model produced by the HC method. The latency of the proposed method is 0.71\% lower compared to HC method.

\begin{figure*}[t]
\centerline{\includegraphics[scale=0.45]{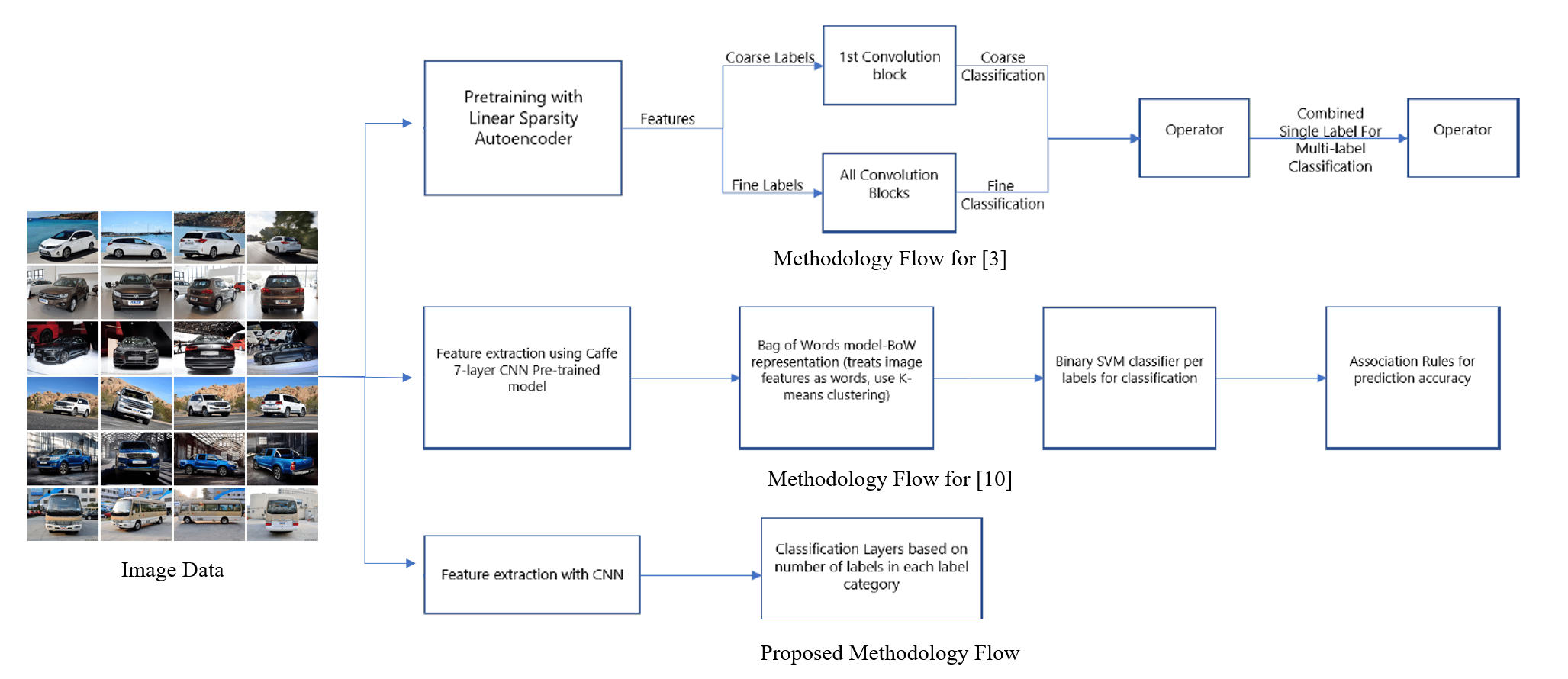}}
\caption{Summary of Comparison With State-of-the-Art}
\label{compare1}
\end{figure*}

Table \ref{tab_goog} shows the comparison between the 3 techniques when applied on GoogleNet. From Table \ref{tab_goog}, it is observed that the proposed approach the same amount of MACC in the trainable layers compared to that of the 2M approach. The proposed approach require about 50\%  less memory and 50\% less latency when compared to the 2M approach. The accuracy and loss values of the 2 approaches are similar and fall within the close regions.

For GoogleNet from Table \ref{tab_goog}, it is seen that the accuracy of the proposed approach lies in close region when compared with the accuracy of the HC technique. The proposed method shows a lower loss, require lower memory size and show require lower latency when compared with the HC method.


Table \ref{tab_squeeze} shows the comparison between the 3 techniques when applied on SqueezeNet. From Table \ref{tab_squeeze}, a consistent result observed that the proposed approach requires 50\% less memory requirement and offers 50\% reduction in latency with the same amount of MACC when compared to the 2M approach. The accuracy and loss values of the 2 approaches are similar and fall within the close regions.

For SqueezeNet from Table \ref{tab_squeeze}, it is seen that the accuracy of the proposed approach lies in close region when compared with the accuracy of the HC technique. The proposed method shows a lower loss, require lower smaller memory and provides a reduction in the latency when compared with the HC method.

This method of multi-label classification shows consistent result when applied to the 3 state-of-the-art CNN models while saving memory size and hardware latency with lower losses.

\section{Comparison with State-of-the-Art}
Some state-of-the-art approaches have been adopted to perform multi-label classification. Kejriwal et.  al \cite{kejriwal2017multi} proposes a Multi Instance Multi Label Classification methodology for Restaurant Images. In this approach, unsupervised learning algorithm (K-means) is used to perform multilabel classification by finding association between Bag of Words (BoW). These BoW are obtained using the feature extraction properties of pretrained CNN models leveraging on transfer learning. This approach requires both supervised and unsupervised  learning to achieve multi-label classification. This approach requires CNN models and unsupervised learning models which will incur more hardware resource compared to our methodology. The process flow for this approach is shown in Fig. \ref{compare1}.

Zhang et. al \cite{zhang2017multi} proposes a multi-label classification model that uses the hidden semantic of different labels. To extract the hidden semantics, an unsupervised learning approach called linear sparsity autoencoder model is used.  Two label sets are distinguished- fine and coarse to enhance classification. A designed operator is used to combine the results from the coarse and fine label classification to give a single to-5 accuracy and top-1 accuracy prediction.  This approach comes with a drawback of pretraining and the employment of 2 models – Autoencoder and CNN model hence it incurs more hardware resources to perform multi-label classification compared to our label. The process flow for this approach is shown in Fig. \ref{compare1}.

\begin{table}[h]
\centering
\caption{Summary of State-of-the-Art Comparison}
\begin{tabular}{|c|c|c|c|}
\hline
\hspace{2mm}Metric  \hspace{2mm}                         & \hspace{2mm}\cite{kejriwal2017multi} \hspace{2mm} & \hspace{2mm}\cite{zhang2017multi} \hspace{2mm}& \hspace{2mm}Ours   \hspace{2mm}   \\ \hline
Employ Only One Model            & x                    & x                  & \checkmark \\ \hline
Employs Only Supervised Learning & x                    & x                  & \checkmark \\ \hline
Performs Multiple Classification & x                    & x                  & \checkmark \\ \hline
Scalable                         & x                     & x                  & \checkmark \\ \hline
\end{tabular}
\label{tab:compare}
\end{table}

\section{Related Work}

Images in real life scenarios most times offer a lot of information. Hence images can be represented by more than one label \cite{wang2016cnn}. Where each label represent mutually exclusive attributes of the image. Multilabel classification addresses the challenge of representing and classifying data samples with more than one label. One limitation in popular conventional classification methods is that they are designed for single labelled data samples \cite{song2018deep}.
Many approaches have adopted deep learning models for multilabel classification.

Shi et. al \cite{shi2017training} proposes a combination of Max-margin, max-correlation objective together with a cross-entropy loss. This method was reported to improve  the accuracy of the multilabel classifier whilst easing threshold determination and the generated model maximized the correlations between the feature vectors. The max-margin objective ensures the minimum score of the positive labels is larger than the maximum score of the negative labels. Max-correlation objective puts the DCNN in a position to maximize the correlations between the learned features and the corresponding ground-truth label vectors in the space. Fully connected layers are thus created for the feature vector and round truth vectors in the projection space and their difference is used to minimize error rates.  

Wang et. al \cite{wang2016cnn} proposes a unified CNN-RNN framework for multi-label image classification. This methodology makes use semantic redundancy and the co-occurrence dependency. The proposed CNN-RNN framework combines the advantages of joint image/label embedding. This approach learns a joint embedding space to bridge the image-label relationship and also the label dependency. The RNN aspect of the framework calculates the probability of a multilabel prediction. During prediction, the multilabel prediction with the highest probability can be approximately found with beam search algorithm.

Song et. al \cite{song2018deep} proposes a deep multi-modal CNN for multi-instance multilabel (MIML) image classification. The model proposed exploits the correlations among class labels by grouping labels in its later convolutional and fully connected layers to learn high level features for related labels within a group. The model represents each image data as a bag of visual instances by exploiting the architecture of CNNs. The model discriminates among visually similar objects belonging to different groups. This approach merges all the group descriptions to generate multi-modal instances. Multi-modal representations combining features from different modalities is utilized to perform multilabel classification.

\section{Conclusion}
This paper introduces a step-by-step and end-to-end methodology of how to design and train a scalable multilabel classifier. This methodology also shows how to expand the capability of a single labelled deep learning model into a multilabel classifier by leveraging on transfer learning with low cost of labelling and achieving impressive accuracy.
Using our methodology, a single model can achieve multiple classification on an image without an increased cost . This method is quite useful for exploiting a single image for various classification.

\bibliographystyle{IEEEtran}
\bibliography{TT.bib}
\end{document}